# Solving MKP Applied to IoT in Smart Grid Using Meta-heuristics Algorithms: A Parallel Processing Perspective

Jandre Albertyn
School of Electrical and Information Engineering
University of the Witwatersrand
Johannesburg, South Africa, 2000
jalbertyn@hotmail.co.za

Ling Cheng
School of Electrical and Information Engineering
University of the Witwatersrand
Johannesburg, South Africa, 2000
ling.cheng@wits.ac.za

Adnan M. Abu-Mahfouz
Council for Scientific and Industrial Research (CSIR)
Pretoria, South Africa, 0184
a.abumahfouz@ieee.org

*Abstract*—Increasing electricity prices in South Africa and the imminent threat of load shedding due to the overloaded power grid has led to a need for Demand Side Management (DSM) devices like smart grids. For smart grids to perform to their peak, their energy management controller (EMC) systems need to be optimized. Current solutions for DSM and optimization of the Multiple Knapsack Problem (MKP) have been investigated in this paper to discover the current state of common DSM models. Solutions from other NP-Hard problems in the form of the iterative Discrete Flower Pollination Algorithm (iDFPA) as well as possible future scalability options in the form of optimization through parallelization have also been suggested.

## I. Introduction

With the introduction of smart home systems and the advancement of the Internet of Things (IoT), new channels of communication are starting to open between customers and their power utility. Smart grids can make use of DSM strategies that enable real time monitoring on the client side of the electrical grid [1]. This can enable utilities to incentivise customers to use appliances at off peak hours by investigating the scheduling, period, and duration at which their customers are consuming electricity. By determining the rate of use of electricity per household, the utility can efficiently price its electric market at the load generation side to set prices for off peak times that are more attractive to the customer. On the client side, DSM systems can monitor price ranges of the electric grid to schedule appliances at specific times to minimise the potential cost that the customer would have to pay towards the utility. It does so by shifting the load of the household to off-peak time slots where able as to fully utilize the discounted price presented by the utility [2]. In doing so, the smart grid enables higher levels of efficiency for the whole system by optimizing its operational capacity. This enables more stability within the electrical system as load is spread more evenly, which in turn reduces the risk of exceeding the capacity within the electrical system, preventing critical failure and potential load shedding.

There are several different methods for DSM that can be found in literature [3] - [6]. A commonality with all the solutions are that they investigate ways in which to solve the Multiple Knapsack Problem by making use of various meta-heuristics to enable the optimisation of scheduling appliances to minimize costs and the Peak to Average Rate (PAR) [7].

The investigation into meta-heuristics used in DSM has proven that there is still further room for optimisation of the MKP. As such, this paper proposes investigating meta-heuristic solutions of another NP-Hard problem that can be closely related to the MKP [8] in the discrete Traveling Salesman Problem (TSP) [9]. Both MKP and TSP share common meta-heuristics and investigating the effects of the iDFPA allows further possibilities for optimisation within the DSM model. This paper also investigates implementing parallelisation strategies to find an even more optimised solution to the MKP problem which in turn will produce more efficient scheduling.

The rest of this paper is organise as follow;. Section II goes into the background describing common heuristic algorithms. Section III investigates existing solutions for smart grid optimization and provides insight into current meta-heuristics and possible improvements. Section IV compares the advantages and disadvantages of the current MPK optimization and presents future research that can be done in the field of meta-heuristics and the MKP.

## II. Background

### A. Flower Pollination Algorithm

The Flower Pollination Algorithm (FPA) developed by Zhou [10] is a nature inspired algorithm that is modelled after the pollination process of plant reproduction. The FPA focuses on optimizing the rate at which flowers are pollinated to enable optimal reproduction. The FPA can be summarised by the following four rules:

- Rule 1 - Biotic and cross-pollination act as global pollination. The agents moving across global space follow the levy-flight distribution.
- Rule 2 - Biotic and self-pollination are forms of local pollination
- Rule 3 - Pollinators can develop a reproduction probability that is proportional to the similarity of two flowers involve
- Rule 4 - Switching between local and global pollination is controlled by a switching probability. This is normally biased slightly towards local pollination

### B. Iterative Discrete Flower Pollination Algorithm

A further optimization of the FPA called the Iterative Discrete Flower Pollination Algorithm (iDFPA) was developed by Strange [11]. The iDFPA is a hybridization of the FPA and the Ant Colony Optimization (ACO) and will be discussed in Section III of this paper.

### C. Genetic Algorithm

Genetic algorithm (GA) introduced by John Holland [12], is an evolutionary search algorithm that is based on genetics and natural selection. GA is applied to find solutions for optimization problems by making use of a sophisticated random search. GA can be classified into four parts. encoding, selection, crossover, and mutation.

The encoding stage allow the chromosomes in the GA to be represented by a binary sting where the digits of string represent the ON or OFF state of the appliance within the context of the DSM. The length of the chromosome can then be used to display the number off appliances that are switched on or off. The population of chromosomes can then be initialised within a period in the smart grid schedule by randomly generating sets of binary strings.

The selection stage works on the survival of the fittest model and compares the current chromosomes to the fitness function determined by the problem space. The selected fittest solutions then enter the crossover state where traits from parent solutions are recombined to form the next generation by selected crossover method. The newly generated solutions then take the place of the weakest solutions in the populations. The mutation stage is finally introduced to add random mutations to the solution of the new generation to attempt to escape the local optimal, before which the process is repeated to find the next generation [13].

*D. Bat Algorithm*

The Bat Algorithm (BA) is a meta-heuristic that was developed by Xin-She Yang by making use of echolocation characteristics in bats as a solution to global optimization problems [14]. The BA can is structured around the following three assumptions:

- Assumption 1: Bats can differentiate between food and obstacles and they use their echolocation to determine how far objects or food are away from the bat.
- Assumption 2: Bats are spawned randomly with a position $x_i$ and velocity $v_i$ along with a fixed frequency and varying wavelength and loudness. The bats can change their wavelength and their pulse emission rate based on how far they are away from their food.
- Assumption 3: Loudness for the bats fall within the range of a constant $A_{min}$ to a maximum $A_0$

The BA makes use of a random uniform distribution along with the frequency range to control the pace and range of the swarming bats while the loudness of each bat acts as a random local search for the swarm. Thus, the BA is a combination of a Particle Swarm Optimization combined with a local search which is controlled by loudness and pulse rate [15].

*E. Teaching-Learning-Based Optimisation*

The Teaching-Learning-Based Optimization (TLBO) was developed by R.V. Rao in 2012 and is based on a school learning process which has been divided into two separate stages [16]. The first stage in the TLBO focuses on teaching. A Gaussian Distribution is used to determine the effect of the teacher on the population of individuals represented by the equation:

$$X_{new} = X_i + r \cdot (X_{teacher} - (T_F \cdot X_{mean}))$$

where $X_i$ is an individual in the population with a vector D to represent all the classes which they are enrolled in. $X_{teacher}$ is the most successful individual in the population according to the problem's fitness function. $r$ is a random distribution with $r \epsilon [1,0]$. $T_F$ is a gaussian distribution $T_F \epsilon [1,2]$ to control the impact of individual quality and $X_{mean}$ is the current mean value of individuals.

The second phase is called the learner phase where each learner can learn from a randomly selected learner in the class $X_{ii}$. If the selected learner $X_{ii}$ is more fit than $X_i$, then $X_i$ moves towards $X_{ii}$ according to

$$X_{new} = X_i + r \cdot (X_{ii} - X_i)$$

otherwise $X_i$ moves away from $X_{ii}$ according to

$$X_{new} = X_i + r \cdot (X_i - X_{ii})$$

If the individual $X_{new}$ has improved, they are deemed as feasible and the algorithm keeps running until it reaches its iteration limit. Additionally, the algorithm needs to deal with infeasible individuals. The following constraints, introduced by Deb [17], are utilized when comparing new individuals:

- The fittest one is chosen between two feasible individuals
- The feasible individual is chosen if on individual is feasible and the other is not
- The individual with the fewest violations is selected if both are infeasible

III. EXISTING SOLUTIONS

*A. Current Smart grid heuristic optimizations*

The IoT has allowed for a wide range of new technologies for communication between the supply of the utility and the demand of the customer. Each home in the smart grid contains an EMC. The EMC enables DSM on the customer side by collecting data from appliances, generation devices and energy storage systems. The EMC then correlates the user preferences with the demand response from the utility to produce an objective function to schedule each appliance to maximize user comfort while minimizing consumer costs and PAR.

In this system, appliances are split into three categories. Fixed appliances, shift-able appliances, and uninterruptable appliances. Fixed appliances cannot be interrupted and are forced to follow the schedule set by the user. Shift-able appliances can be scheduled and interrupted at any time and uninterruptable appliances can be set to start at any scheduled time, but they cannot be interrupted once they have started.

The EMC makes use of meta-heuristic algorithms to optimize an NP-hard MKP. The items in the MKP are represented by the appliances, the energy consumed by each appliance considered to be its weight and the operational cost would be its profit.

Consumers are then able to interface with the system and provide the EMC with appliances that they want to have scheduled, how long they want them to be scheduled for and other user comfort options. The EMC will then use its meta-heuristic optimization to generate a schedule and shift the load accordingly by switching appliances on/off at the appropriate times.

Rahim [3] uses the most common method found in literature in the form of an improved ACO meta-heuristic to optimize the MKP. They were able to conclude that there was a decrease in operational expenses as well as a reduced PAR.

Khalid [4] also followed the same EMC model to construct an MKP optimisation, but incorporated the BA, the FPA and the Hybrid Flower Pollination Bat Algorithm (HFBA), which is a hybrid of the BA and the FPA. The HFBA follows the same structure as the BA but replaces the random generation step with the flower swarm step from the FPA. All three algorithms outperformed the non-optimised solution in terms of cost and PAR minimization as well as user comfort. A further comparison was done between the three algorithms and it was found that the HFBA outperformed both other algorithms over three case studies.

Iqbal [5] further explores the DSM Smart grid model by comparing a set of existing optimization techniques with proposed hybrid optimization techniques namely, Genetic Teaching Learning Based Optimization (GTLBO), Flower Pollination Teaching Learning Based Optimization (FTLBO), Flower Pollination Bat Algorithm (FBAT) and Flower Pollination Genetic Algorithm (FGA)

- GTLBO is a hybrid optimization algorithm that combines elements of the GA with that of the TLBO. This is done to attempt to harness the strengths of both algorithms as the TLBO performs well in exploitation mode, but it is ineffective at performing global searches. Inversely, the GA is incredibly well suited for global searches without being trapped in local optima. The techniques are combined by using the TLBO during the initial step, then applying the mutation operator of the GA after updating the population in learner mode.
- FTLBO is the hybrid combination of the FPA and TLBO techniques. The FTLBO still follows the initial steps of the TLBO. A change occurs in the teaching and learning phase, these stages are replaced by searching for random flowers based off the FPA. The fitness value is then compared to the current best solution to determine if the new solution is to be used as the global best solution
- The FBAT used by Iqbal is hybridized in the exact same way as the HFBA optimization that was constructed by Khalid. Both reports also confirm that the FBAT outperforms FPA and BAT in cost and PAR as well as user discomfort levels.
- FGA is the combined hybridization of the FPA and the GA. FGA still follows the exact same step as in GA, but the crossover and mutation steps are replaced by searching for random flowers in the neighbourhood of FPA.

Iqbal found that the new proposed hybrid algorithms outperformed their existing counterparts in terms of reduced cost, reduced PAR, and reduced user discomfort. It was also noted that there was a trade-off between costs and user discomfort. This is due to the fact the lowest costs would move most of the appliance load to off peak hours whereas the customer prefers to run their applications at peak hours.

*B. Iterative Discrete Flower Pollination Algorithm*

The iDFPA developed by Strange [7] describes the formulation of a meta-heuristic algorithm used to solve the discrete space NP-Hard Traveling Salesman Problem. Since the FPA is used to solve for problems in continuous space, it was adapted and combined with the ACO algorithm to form the DFPA. DFPA hybridizes the ACO algorithm and the FPA by making use of the local and global search elements from the FPA and the multi agent elements from the ACO.

The algorithm requires a distance matrix $D$ to be constructed as a $n \times n$ matrix where $n$ is the number of nodes for the given problem. The $D$ matrix is then populated with a diagonal of zeroes and distance values $d_{ij}$ to represent the distance between the i[th] and j[th] nodes. For a symmetric TSP the distance from node $i$ to node $j$ will always be the same as the distance from node $j$ to node $i$. An $n \times n$ cost matrix $C$ then produces as an inverse of the distance matrix to determine the probability for each path.

The algorithm uses $m$ agents simultaneously to find the best solution. It does this by forming a Markov Chain by using probabilities determined from the current node. The algorithm also makes use of a switching probability $\rho \in (0, 1)$ to switch between local and global search.

Each $m$ agent adds its tour to the set of tours $S = \{s_1, s_2, ..., s_m\}$ where $s_i = (s_i^1, s_i^2, ..., s_i^n)$ is an ordered sequence tour which is a permutation of $V = \{v_1, v_2, ..., v_n\}$. This enables the use of $V' = V$ where $V'$ is a temporary set to represent the nodes still left within the tour. Next, randomly select $s_i^1$ from the list of nodes in $V'$ and set $v_{prev} = s_i^1$ and remove it from $V'$ so that $V' = V' - \{s_i^1\}$. To determine the rest of the tour, first generate the normalization constant $A = \sum_{\forall v_k \in V'} c_{ik}$. The probability of each node in $V'$ is then given by

$$p_{ij} = \frac{c_{ij}}{A}$$

A random number $r \sim U(0,1)$ is generated. If $r > \rho$ then the global search algorithm is used to determine $s_i^j$ else the local search algorithm is used. Finally, the j[th] element is removed from the temporary tour variable so that $V' = V' - \{s_i^j\}$ and $v_{prev} = s_i^j$ after which the $j$ variable is incremented to get the next element in the tour. Once all the tours in S have been determined, the minimum distance is chosen as the solution.

For local search within DPFA, a search subset $V'' = \{v | d(v, v_{prev}) \leq r_{dist}, v \in \bar{V}\}$ can be used, where $V''$ is the ordered probability set of $V'$. The global search is used if there are no more nodes in the radial cluster.

The global search makes use of a discrete Levi Distribution Function $L(\bar{V})$ which is defined as

$$f(v_i; \bar{V}) = \Pr(v_i)$$
$$= F_L(\Phi \sum_{k=1}^{i} p_k) - F_L(\Phi \sum_{k=1}^{i-1} p_k)$$

where $F_L$ is the Levy CDF and $\Phi$ is a constant. After the discrete Levy Distribution is applied to the node $L(\bar{V})$, the next node $s_i^j \sim L(\bar{V})$ is generated using

$$l^* = \min \{l \in \{1,2, \cdots, |\bar{V}|\} : (\sum_{i=1}^{l} p_i) - r \geq 0\}$$

Now that the DFPA can escape local optima with the FPA and can incorporates multiple agents, it is able to focus both on exploration and exploitation. The algorithm can still be improved however, by introducing memory into the system. The iterative Discrete Flower Pollination Algorithm (iDFPA) was created by adding both the best tour update and the rejection update to the DFPA process. This adds some extra complexity to the algorithm, but further increases the convergence rate.

The first element that is adjusted in the system is the cost matrix. The new cost matrix $C^t$ represents the cost of the cost $c_{ij}^t$ to travel between each node $v_i$ and $v_j$ during the $t^{th}$ iteration of the process. Evaporation follows the formula

$$c_{ij}^{t+1} = (1 - \alpha)c_{ij}^t$$

where $\alpha$ controls the rate of evaporation. This is to allow for bad memory to be filtered out of the system. This then allows for use of the filtering processes known as the Best Tour Update (BTU). The best tour update allows for the cost matrix to be updated, but only if the path was travelled during the minimum distance tour $s^*$. The cost matrix is the updated as follows if $arc(i,j)$ is part of $s^*$;

$$c_{ij}^{t+1} = c_{ij}^t + \gamma \frac{d_{ij}}{d_{s^*}}$$

where the constant $\gamma$ controls the magnitude of the best tour update.

The last part of the iDPFA is the Rejection Update (RU). The RU also utilizes the tour distance of the iDPFA. If a new tour distance is introduced, it is automatically added to the list of accepted tours $S' = \{s_1', s_2', \cdots\}$. If the tour is not shorter than the minimum tour, exponential annealing is introduced by

$$T_{curr} = e^{-\omega \frac{1}{q} \frac{N_{curr}}{N}}$$

where $\omega$ is a constant that controls the utilised region of the exponential function and $q$ controls its' shape. $N_{curr}$ is the current iteration number and $T_{curr}$ is the annealing values corresponding to the current iteration. Tour distances that are longer than the shortest tour distance then have a probability of being accepted of

$$p = e^{\frac{-\triangle dist}{(T_k \times d_{prev})}}$$

where $\triangle\, dist = d_{sj} - d_{prev}$. A random number $r \sim U(0,1)$ is then compared to $p$. If $p > r$ then the solution is accepted otherwise it is rejected. The rejection cost update is the shown as follows:

$$c_{ij}^{t+1} = c_{ij}^t + \beta \sum_{\substack{\forall s_k' \in S' \\ containin \\ arc(i,j)}} \frac{1}{d_k'}$$

where $d_k' \in d'$ and $\beta$ is a constant that controls the magnitude of the Rejection Update on the cost matrix.

*C. Parallelisation of the travelling salesman problem*

All the processes investigated above made use of single processor, sequential methods to solve for NP-hard problems. In [18] Randall investigated different parallel ACO architectures and compared the efficiency and speedup of parallelised ACO compared to serial implementation on the same computer. Randall did also note that large overhead in communication may severely hinder the performance of these parallel systems. As such all the implementations were assumed to run on a distributed network.

Randall's paper focused on five different parallel configurations

*1) Parallel Independent Ant Colonies*

Colonies are run sequentially over different processors. Each colony has its own set of key parameters that differ from other colonies. These processes can run independently from each other.

*2) Parallel Interacting Ant Colonies*

This process follows the same steps as the method above, but at the end of each iteration the best pheromone levels are communicated to the rest of the colonies. This can be costly as large pheromone structures will take time to broadcast.

*3) Parallel Ants*

Several ants are clustered onto each processor. The master processor keeps track of all user defined variables, ant starting points and the global pheromone levels and is responsible for the final solution. The largest overhead is each ant sending back its pheromone structure to the master.

*4) Parallel Evaluation of Solution Elements*

Each ant does its own processing and determines its own best solution by being able to evaluate all the solution elements. Since each ant is doing its own computations, they can run independently. The process can be parallelised between each step.

*5) Parallel Combination of Ants and Evaluation of Solution Elements*

This is a combination of Parallel Ants and Parallel Evaluation of Solution Elements. All the ants are divided equally into groups where the groups process their own solution elements. This would generally require a much larger number of processors.

The paper implemented Parallel Ants since the cost of interacting ants becomes too large due to the overhead of transferring the pheromone structure. Parallel Evaluation requires a lot of processors and would be more useful for very complex evaluations. The Travelling salesman problem does not have an extremely high level of complexity so this would not need to be implemented.

Speedup was calculated by dividing the processing time of serial code by the wall clock time of the parallel code. Efficiency was then calculated by speedup divided by number of processors. It was found that speedup and efficiency are only effectively achieved for problems that used more than 300 nodes.

Chen expanded further on the parallel ant network in [19]. Chen made use of MPP architecture which allows a group of ants to be clustered on each processor to implement a parallel ACO system (PACO). This allows a massively parallelised solution as each group can find their own solution. MPP has the benefit of high bandwidth communication between each processor as well. To make sure that processors do not reach their local minimum, information is exchanged between processors.

This exchange of information is done in terms of sharing the best solution per processor. Each MPP processor is assigned a time interval when starting its ACO. After this interval is exceeded, the processor looks for its own partner to communicate with. This is done by finding the partner that shares the least amount of common edges with the current solution. After a partner has been found, a global pheromone update is committed to the current processor and a new period is calculated based on the amount of common edges shared with the chosen partner. The processor will then again work with local pheromone updates to determine the best solution and calculate a new period for communication. The more similarity between two partners, would imply that there will be a faster convergence and as such a shorter period is needed before the next exchange interval.

## IV. COMPARISON AND DISCUSSION

### A. Comparison

GTLBO, FTLBO, FBAT and FGA were compared to their parent meta-heuristics in Table 1 based on three case studies done in [5]. The hybridised meta-heuristics that were developed were generally able to outperform their parents. Trade-offs between the meta-heuristics were also introduced to explain why different levels of performance could be expected for each meta-heuristic.

### B. Discussion

Load shedding is an ever present risk in South Africa and as such, it is important for us to start looking at effective ways in which we are able to preserve power to ensure that we do not overload an already taxed system. We have however also been able to vastly increase our ability to keep track of various systems in and around our homes by the technologies provided through the IoT. One of the devices that have been enabled through this technology is the smart grid and its EMC.

Each of the papers that were investigated in terms of the smart grid systems and their optimization has shown the importance of meta-heuristics in a practical sense. On average each existing algorithm was able to significantly reduce the daily cost of running electrical applications in a SG.

Rahim [3], Khalid [4], Iqbal [5] and Hussain's [6] articles showed that on average, the existing meta-heuristics can reduce daily electrical cost by a significant amount of about 10-20% just by making sure that they are scheduled by a basic optimization heuristic. The development of more complex meta-heuristics has been able to boost the reduction in cost to between 40% and 65%. Iqbal's article also shows that this significant increase in cost reduction by using more sophisticated meta-heuristics had also helped to reduce the PAR within smart grids while providing similar user comfort levels.

The examples in the literature shows just how effective well optimized meta-heuristic solutions can work for real world applications as shown in the MKP for Smart grid DSM. Every improvement in meta-heuristics enables a more optimized solution leading to more efficient systems. Further potential optimizations can be seen in the TSP domain as shown in Table 1.

This is largely due to resulting meta-heuristics being able to eliminate weaknesses in their parent algorithms. This however comes at a trade-off by introducing extra complexity into the meta-heuristic. The two most successful examples of this is the FGA and FTLBO. Both algorithms enabled faster convergence for the PSO based algorithms which enabled a reduced PAR along with a cost reduction due fast conversion along with the possibility of escaping local optima.

The iDPFA developed by Strange has shown a significant increase in performance over the GA and ACO algorithms for a wide range of problems. This ACO algorithm that has already shown a lot of promise in the MKP problem space and has also been further improved by making use of parallelisation. This improvement has only been proven to be successful in a problem space with more than 300 nodes and as such is not feasible for small scale optimization problems. The parallel framework is however highly scalable and can solve big maps with more accuracy and efficiency than a single concurrent algorithm.

| Algorithm | Trade-offs | | Optimization Metrics [5] | | |
|---|---|---|---|---|---|
| | *Advantages* | *Disadvantages* | *Cost Reduction* | *PAR* | *User Discomfort* |
| GA | Optimal at performing global searches and escaping local optima. Converges quickly resulting in high cost reduction. | Poor performance for local searches. Unable to reduce PAR. Lowering cost while not lowering PAR leads to high user discomfort | High | Low | High |
| TLBO | Optimal at performing local searches. Faster convergence than PSO based algorithms | Poor performance for global searches. Unable to reduce PAR, resulting in high user discomfort | Moderate | Low | High |
| FPA | Switchover criteria enables the algorithm to escape local optima whilst staying effective in global searches | Poor cost and PAR reduction, possibly PSO algorithms can get trapped in local optima when solving complex multimodal problems. Slow convergence [20] | Low | Low | Low |
| BA | Combines local search step with PSO to linit weak local search | Same issue as FPA | Low | Low | Low |
| GTLBO | Combines both GA and TLBO to take advantage of exploitation and eploration. | Non-proportional reduction in decrease of cost lead compated to PAR leads to user discomfort. Futher optimisation is required | High | High | Low |
| FTLBO | Combined switchover and fast convergence. TLBO algorithm eliminates less solutions possible solutions than GA, allows for higher PAR | Slower convergence than GTLBO. Higher cost reduction leads to higher user discomfort. | Very High | Very High | Moderate |
| FBAT | Faster convergence due to flower step instead of random step of the BA | Still suffers from slow convergence due to lack of elimination. | Moderate | Moderate | Moderate |
| FGA | Faster convergence than FPA due to elimination criteria. FPA global search capibility enables the algorithm to escape local optima | Non-proportional reduction in decrease of cost lead compated to PAR leads to user discomfort. Futher optimisation is required | Extremely High | High | High |

TABLE 1.    COMPARISON OF HEURISTIC TECHNIQUES

*C. Future research direction*

ACO and FPA Hybrid algorithms have been tested on the MKP problem for smart grids and both have shown promising results. The FPA ACO Hybrid of iDPFA that was created by Strange has shown to outperform the ACO for the TSP discrete NP-Hard problem. Implementing the iDPFA on the MKP problem should be investigated further to see if it can outperform the other FPA hybrid optimizations as well as the ACO

Another future project should be to incorporate the PACO algorithm into the iDPFA meta-heuristic and to test it on a large-scale discrete NP-Hard problem.

## V.    CONCLUSION

Smart grid applications have several NP-hard problems such as MKP. The Demand side management system of the smart grid has shown its need for an optimization solution that can correctly schedule various appliances to minimize operating cost, PAR and to maximize user comfort.

Current solutions for smart grid optimization have also shown the need for more advanced, improved meta-heuristic algorithms as shown by the by the decrease in costs by both the ACO and the hybrid FPA algorithms. The iDFPA has been suggested to fill this need

Lastly there is still a lot of utilization that is required in the field of parallelisation of discrete algorithms. This will aid in any future scaling problems as well as solving for NP-Hard problems from the supply side for load distribution problems.

## REFERENCES

[1] Logenthiran, T., Srinivasan, D. and Shun, T.Z., "Demand side management in smart grid using heuristic optimization". *IEEE transactions on smart grid*, 3(3), pp.1244-1252, 2012.

[2] Malaye, D., Dange, K., Barhate, T., Narwade, S., Kulkarni, R.D., Vadirajacharya, K. and Thorat, S., "Current trends of implementing smart grid for enhancing the reliability of power utility network." in *2017 2nd International Conference on Communication Systems, Computing and IT Applications (CSCITA)*, pp. 281-285, April 2017, IEEE

[3] Rahim, S., Iqbal, Z., Shaheen, N., Khan, Z.A., Qasim, U., Khan, S.A. and Javaid, N., "Ant colony optimization based energy management controller for smart grid," in *2016 IEEE 30th International Conference on Advanced Information Networking and Applications (AINA)* pp. 1154-1159, March 2016, IEEE.

[4] Khalid, R., Javaid, N., Rahim, M.H., Aslam, S. and Sher, A., "Fuzzy energy management controller and scheduler for smart homes," *Sustainable Computing: Informatics and Systems*, 21, pp.103-118, 2019.

[5] Iqbal, Z., Javaid, N., Mohsin, S.M., Akber, S.M.A., Afzal, M.K. and Ishmanov, F., "Performance analysis of hybridization of heuristic techniques for residential load scheduling," *Energies*, 11(10), p.2861, 2018.

[6] Hussain, H.M., Javaid, N., Iqbal, S., Hasan, Q.U., Aurangzeb, K. and Alhussein, M., "An efficient demand side management system with a new optimized home energy management controller in smart grid," *Energies*, 11(1), p.190, 2018.

[7] Liu, Y., Yuen, C., Huang, S., Hassan, N.U., Wang, X. and Xie, S., "Peak-to-average ratio constrained demand-side management with consumer's preference in residential smart grid," *IEEE Journal of Selected Topics in Signal Processing*, 8(6), pp.1084-1097, 2014

[8] Chekuri, C. and Khanna, S., "A polynomial time approximation scheme for the multiple knapsack problem," *SIAM Journal on Computing*, 35(3), pp.713-728, 2005.

[9] Maredia, A. and Pepper, R., "History, Analysis, and Implementation of Traveling Salesman Problem (TSP) and Related Problems," *Department of Computer and Mathematical Sciences*, University of Houston-Downtown, 2010

[10] X.-S. Yang, M. Karamanoglu, and X. He, "Flower pollination algorithm: a novel approach for multiobjective optimization," *Engineering Optimization*, vol. 46, no. 9, pp. 1222–1237, 2014.

[11] R. Strange, A. Y. Yang and L. Cheng, "Discrete Flower Pollination Algorithm for Solving the Symmetric Travelling Salesman Problem," *2019 IEEE Symposium Series on Computational Intelligence (SSCI)*, Xiamen, China, 2019, pp. 2130-2137

[12] Mitchell, M., "An introduction to genetic algorithms," MIT press, 1998.

[13] Deng, Yong, Yang Liu, and Deyun Zhou. "An Improved Genetic Algorithm with Initial Population Strategy for Symmetric TSP," *Mathematical Problems in Engineering* 2015

[14] Yang, X.S. "A new meta-heuristic bat-inspired algorithm", *Nature Inspired Cooperative Strategies for Optimization*, Springer: Berlin/Heidelberg, Germany, 2010; pp. 65–74.

[15] Kongkaew, W., "Bat algorithm in discrete optimization: A review of recent applications," *Songklanakarin Journal of Science & Technology*, 39(5), 2017

[16] Črepinšek, M., Liu, S.H. and Mernik, L., "A note on teaching–learning-based optimization algorithm," *Information Sciences*, 212, pp.79-93, 2012.

[17] Deb, K. and Agrawal, R.B., "Simulated binary crossover for continuous search space," *Complex systems*, 9(2), pp.115-148, 1995.

[18] Randall, M. and Lewis, A., "A parallel implementation of ant colony optimization," *Journal of Parallel and Distributed Computing*, 62(9), pp.1421-1432, 2002.

[19] Chen, L., Sun, H.Y. and Wang, S., "A parallel ant colony algorithm on massively parallel processors and its convergence analysis for the travelling salesman problem," *Information Sciences*, 199, pp.31-42, 2012.

[20] Beheshti, Z. and Shamsuddin, S.M.H., "A review of population-based meta-heuristic algorithms," *Int. J. Adv. Soft Comput. Appl*, 5(1), pp.1-35, 2013.